\pdfoutput=1

\documentclass[11pt]{article}

\usepackage{emnlp2021}

\usepackage{times}
\usepackage{latexsym}
\usepackage{xspace}
\usepackage[hang,flushmargin]{footmisc}

\usepackage[T1]{fontenc}

\usepackage[utf8]{inputenc}

\usepackage{microtype}

\usepackage{graphicx}
\usepackage{arabtex}
\usepackage[utf8]{inputenc}
\usepackage{utf8}
\usepackage{url}
\newcommand{\Ar}[1]{{\small \<#1>\xspace}}

\newcommand{\TrAr}[1]{\arabtrue\transfalse {\scriptsize \Ar{#1}} /\arabfalse\transtrue \RL{#1}\arabtrue\transfalse} 

\newcommand{\Tr}[1]{\arabtrue\transfalse \arabfalse\transtrue \RL{#1}\arabtrue\transfalse} 

\newcommand{\corpora}{\cal{L\^{i}sa\=n}\xspace}
\newcommand{\arcorpora}{\Ar{لِسَان}\xspace}
\newcommand{\adat}{\cal{ADAT}\xspace} 

\title{\corpora: Yemeni, Iraqi, Libyan, and Sudanese Arabic Dialect Corpora with Morphological Annotations}

\author{Mustafa Jarrar$^1$ \\
  \texttt{mjarrar@birzeit.edu} \\\And
  Fadi A. Zaraket$^2$ \\
  \texttt{fz11@aub.edu.lb~~}\\ \And
  Tymaa Hammouda$^1$ \\
  \texttt{~~1171779@student.birzeit.edu} \\ \AND
  Daanish Masood Alavi$^3$ \\
  \texttt{masoodd@un.org} \\\And
  Martin W\"{a}hlisch$^3$ \\
  \texttt{waehlisch@un.org} \\
} 

\newcommand\blfootnote[1]{%
  \begingroup
  \renewcommand\thefootnote{}\footnote{#1}%
  \addtocounter{footnote}{-1}%
  \endgroup
}

\begin{document}
\setcode{utf8}

\maketitle

\begin{abstract}
This article presents morphologically-annotated 
Yemeni, Sudanese, Iraqi, and Libyan Arabic dialects 
(\corpora) corpora. 
\corpora features around 1.2 million tokens. 
We collected the content of the corpora from several 
social media platforms. 
The Yemeni corpus (\~ 1.05M tokens) was collected automatically from Twitter. 
The corpora of the other three dialects 
(~\ 50K tokens each) 
came manually from Facebook and YouTube posts and 
comments.

Thirty five (35) annotators 
who are native speakers of the target dialects
carried out the annotations. 
The annotators 
segemented all words in the four corpora 
into {\em prefixes}, {\em stems} and {\em suffixes} 
and labeled each 
with different morphological features such as 
{\em part of speech}, {\em lemma}, and a {\em gloss} 
in English. 
An Arabic Dialect Annotation Toolkit (\adat) was 
developped for the purpose of the annation. 
The annotators were trained on a set of guidelines 
and on how to use \adat. 
We developed \adat to assist the annotators and 
to ensure compatibility with SAMA and Curras tagsets. 
The tool is open source, and the four corpora are 
also available \href{https://portal.sina.birzeit.edu/curras}{online}. 
 
\end{list} 
\end{abstract}

\section{Introduction}
\label{sec:intro}

\blfootnote{
  $^1$~University of Birzeit, Birzeit, Palestine \\
  $^2$~American University of Beirut, Beirut, Lebanon \\
  $^3$~Department of Political and Peacebuilding 
  Affairs, United Nations, New York \\ 
} 

Around 300 million people in 23 countries speak and 
use the Arabic language in their daily lives. 
Classical Arabic (CA) is the old form of Arabic 
that is used in historical texts. 
Modern Standard Arabic (MSA) is used in 
formal communications including 
newspapers, media outlets, 
educational material and most of the televised 
content. 
Dialectal Arabic (DA) appears in colloquial and 
informal day-to-day communications. 
DA volume is massively increasing on social media. 
Processing and understanding such content in 
natural language processing (NLP) tasks is 
challenging~\cite{DH21}. 
This is mostly because Arabic dialects are 
under-resourced and have no standard orthography.

DA and MSA differ in four major ways: 

(i) \textbf{Phonology}: 
People pronounce words differently with varied 
intonation and stress, and write them as pronounced. 
For example, the letter (\TrAr{ق}) is pronounced as 
'g' in Libyan and Sudanese and more of a 'k' in 
Yemeni as in the word 
\Ar{قال}/\textit{kal} (say). 
Iraqi switches between 'q' and 'k' as in 
\TrAr{قال} and \Ar{وقت}/\textit{wakt} (time). 
The letter (\Ar{ث}) typically denoting the sound 'th' 
in MSA, is pronounced as 's' in 
Sudanese and more of a 't' in Libyan as 
in the word \TrAr{ثياب}(cloth). 
The letter (\Ar{كـ}) typically denoting 
a 'K' sound in MSA, is pronounced \Tr{تش} 
in Iraqi as in the word \Ar{كلب}/\Tr{تشلب} (dog). 

(ii) \textbf{Morphology}: 
Arabic dialects are similar to MSA in 
inheriting templatic morphology where affixes
play an important role. 
However, major differences exist between dialects 
and MSA and among dialects themselves. 
For example, negating the verb \TrAr{أكلت} (I ate) 
in MSA precedes it with the particle \TrAr{لَم}
and inflects the verb itself to produce
become \TrAr{لم آكل}. 
The Lybian dialect uses the prefix \TrAr{م} 
and the suffix \TrAr{تش} for negation, 
as in \TrAr{مأكلتش} (I did not eat); 
while the prefix \TrAr{ما} is enough in Sudanese. 

In \textcolor{orange}{Yemeni}, the prefix (\TrAr{لـ}) 
is used to negate the imperative as in 
\TrAr{لتخافون}(Do not be afraid), which 
is a short of the \TrAr{لا} particle in MSA, 
as in \TrAr{لا تخافوا}. The \TrAr{مو} and 
\TrAr{مش} particles are also common replacements for 
the MSA negation particles such as \TrAr{ليس}. 
As will be discussed in section 
\ref{sec:Guidelines}, many of the MSA particles 
are used as affixes in Arabic dialects.

(iii) \textbf{Lexicon}: 
Each dialect has its own unique lexicon entries 
that are not used in MSA or other dialects. 
The word \TrAr{عنجاص} (plum) is an Iraqi variant 
of the MSA \TrAr{إجّاص} (pear), while Iraqi uses 
\TrAr{عرموط} to denote pears. 
The Sudanese uses \TrAr{زول} to denote \TrAr{رجل} (man).
These variations can prove embarrassing as \TrAr{تينة} 
(tree of figs) in MSA denotes human body's bottom in 
Libyan. 
Yemeni uses \TrAr{عشار} and \TrAr{قطيب} 
for \TrAr{مخلل} (pickles) 
and \TrAr{زبادي} (thick yogurt), respectively. 
It shares with Iraqi \TrAr{ميز} and \TrAr{بنكة} for 
MSA's \TrAr{طاولة} (table) and \TrAr{مروحة} (fan), 
respectively. 
 
\subsection{Contributions}
This paper contributes to addressing the problem of 
under-resourced Arabic dialects and presents 
\corpora ~\arcorpora `which consists of 
four morphologically annotated corpora of 
the Yemeni, Iraqi, Sudanese and Libyan Arabic dialects.
We collected the text of the corpora from Facebook, 
Twitter, and YouTube. 
We then tokenized and manually annotated the text 
with morphological attributes. 
The annotation methodology we followed is similar to 
that advised for annotating Palestinian and 
Lebanese dialects \cite{JHRAZ17,EJHZ22}, and 
based on SAMA tagsets \cite{MaamouriSama2010}. 
To support and streamline the annotation process, 
we developed the Arabic Dialect Annotation Toolkit 
(\adat) which we also provide as an open source 
online contribution.

\corpora consists of about 1.2 Million fully-annotated 
tokens: Yemeni (1.1M), Iraqi (46K), Libyan (52K), and 
Sudanese (52K). 
Thirty five annotators helped annotate the corpora. 
The \corpora corpora and \adat are both available 
at (\url{https://portal.sina.birzeit.edu/curras}) 
and are open source for academic research purposes.

The rest of the paper is organized as follows. 
Section ~\ref{sec:related} overviews the related work, 
Section ~\ref{sec:collection} describes the 
corpora collection process, and 
Section ~\ref{sec:method} presents the 
annotation methodology. 
We conclude and discuss future directions in 
Section~\ref{sec:conclusion}.

\section{Related Work}
\label{sec:related}
This section reviews efforts to create annotated 
corpora for Arabic dialects.

An early Treebank~\cite{maamouri-etal-2006-developing} 
was created for the Jordanian dialect. 
A Palestinian dialectal corpus, 
called Curras~\cite{JHRAZ17,JHAZ14} and consists 
of 56K tokens, is a more recent corpus collected from 
Facebook and scripts of the Palestinian series 
``Watan Aa Watar''. 
Each word in the corpus was then manually annotated 
with a set of morphological attributes. 
Curras was recently revised and extended with a 
Lebanese corpus (10k Tokens) to form a more 
Levantine corpus \cite{EJHZ22}.

The Egyptian Arabic corpus 
CALLHOME ~\cite{CallHome_1997} consisted of 
transliterations of telephone conversations in 
Egyptian. 
ECAL~\cite{EcalKilany_2002} built on CALLHOME to 
provide morphological analysis of the Egyptian dialect. 
An extension of ECAL was 
CALIMA ~\cite{habash-etal-2012-morphological}.  
The COLABA project ~\cite{Colaba_diab_2010} gathered 
resources in dialectal Arabic from online blogs. 
This combination of projects gradually led 
to constructing the Egyptian TreeBank 
(ARZATB) \cite{maamouri-etal-2014-developing}.

Other efforts to create morphologically annotated 
corpora follow.
\citet{SevenCorpora} presented a 200K tokens corpus  
for seven different Arabic dialects including 
Taizi (Yemen), Sanaani (Yemen), Najdi (KSA), 
Jordanian, Syrian, Iraqi and Moroccan. 
MADAR~\cite{bouamor-etal-2014-multidialectal} 
is an ongoing multi-dialect corpora covering 
26 different cities and their corresponding dialects. 
The work in \citet{TunisianCorpus} presented the first 
release of an 
Arabizi Tunisian corpus (42K tokens). 
The GUMAR Emirati dialect corpus consists of about 
200K tokens collected from Emirati 
novels \cite{khalifa-etal-2018-morphologically}.

Two NLP competition tasks on 
{\em nuanced Arabic dialect identification} (NADI) 
in 2021 \cite{abdul-mageed-etal-2021-nadi} and 2019 
~\cite{abdul-mageed-etal-2020-nadi} provided 
researchers with Arabic dialect data from 21 countries. 
NADI targeted the identification of 100 different 
province level dialects in 21 Arab countries. 
They provided competitors with 21,000 tweets  labeled 
with a province level dialect in addition to a 
10 million tweet dataset with no labels. 

NADI followed the 
{\em fine grained Arabic dialect identification} task
~\cite{salameh-etal-2018-fine} that targeted 
identifying up to 25 city level dialect variations in 
addition to MSA.  
The task provided two corpora: 
(i) the first is composed of 10,000 
{\em basic travel expsression corpus}(BTEC)~\cite{kageura-kikui-2006-self} 
sentences translated to the dialects of five main 
cities, and a separate set of 2,000 BTEC sentences
 translated to 25 city dialects.

\section{Corpus Collection}
\label{sec:collection}
We collected \corpora  from social media networks, mainly from Twitter, Facebook, and YouTube.
Three corpora (Iraqi, Libyan, and Sudanese) were collected manually, 
while the Yemeni corpus was collected automatically. 
The manual collection was carried out by native speakers who carefully 
selected posts and comments discussing politics and general affairs. 
We required the selected comments and posts to be at 
least 10 words and not larger than 30 words. 
We also required them to have at least one 
colloquial word belonging to the target dialect. 

The Yemeni corpus was collected through the Twitter 
API using keywords related 
to the current political situation in Yemen. 

To ensure that each of the collected tweets 
contained at least one colloquial Yemeni word, 
we filtered the tweets using a list of typical and 
distinctive colloquial Yemeni words. 
No specific sub-dialect was preferred in any of 
our four dialect corpora as we aimed to develop 
a general corpus for each dialect. 

Table~\ref{t:basic_stat} provides general 
statistics about each corpus.

\begin{table}[tb!]
\centering
\begin{tabular}{lrr}
\hline
\textbf{Corpus} & \textbf{Tokens} & \textbf{Documents}\\
\hline
Yemeni & 1,098,222 & 38,819 Tweets \\ 
Iraqi & 45,881 & 3,326 Threads \\ 
Libyan & 51,686 & 3,053 Threads \\ 
Sudanese & 52,616 & 3,000 Threads \\ \hline
Total & \textbf{1,248,405}& \textbf{48,198} \\ \hline
\end{tabular}
\caption{Number of documents and tokens per corpus} 
\label{t:basic_stat} 
\end{table}

\section{Corpus Annotation Methodology}
\label{sec:method}
\begin{figure}[!t] 
\centering
\includegraphics[width=0.35\paperwidth]{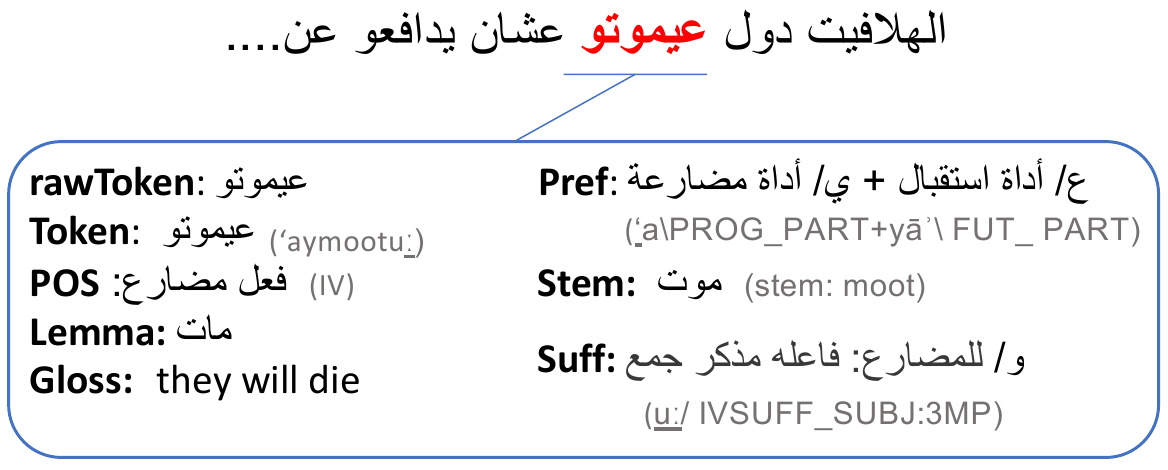}
\caption{Example of an annotated token in context}
 \label{fig:annotation_example}
\end{figure}

This section presents the approach we used to 
annotate our four corpora. 
First, we define the tags we used in the annotation; 
then we describe the tool and the methodology used 
to annotate each word in context. 
Figure~\ref{fig:annotation_example} shows 
an example Sudanese phrase with the word 
\TrAr{عيموتو} (they will die) 
 and
its POS, Prefix, Suffix, Stem, Lemma and Gloss
annotations.

\subsection{Annotation Framework}
Based on the annotation framework used to annotate 
the Curras and Baladi corpora 
~\cite{JHRAZ17,EJHZ22}  
as well as the framework found in
 ~\cite{MaamouriSama2010}, 
we define our annotation framework of a token as a tuple:\\

\noindent{$\langle$ \textit{rawToken},\textit{Token}, \textit{Prefixes}, \textit{Stem}, \textit{Suffixes}, \textit{POS}, \textit{Lemma}, \textit{Gloss} $\rangle$.} \\

Where
\textit{rawToken} is the raw word as it appears in 
the corpus; 
\textit{Token} is a normalized version of 
\textit{rawToken}; and
\textit{Prefixes}, \textit{Stem}, and \textit{Suffixes} 
form the segmentation of the \textit{Token}. 
\textit{POS} is the part-of-speech and 
\textit{Lemma} is the lexicon conical form of the \textit{Stem}. 
The gloss is the meaning of the \textit{Token} 
in English. 
The \textit{Prefixes} and \textit{Suffixes} 
use the '+' separator to separate parts from 
\textit{Token} and the '/' separator to separate
the prefix and suffix POS labels, respectively. 

To ensure maximum compatibility with other MSA 
and dialectal corpora, 
we adopted the morphology tags used in 
LDC’s BAMA and SAMA databases~\cite{MaamouriSama2010}, 
which are commonly used for annotating Arabic corpora 
such as the Palestinian Curras~\cite{JHRAZ17}, 
the Lebanese Baladi, and the 
Emirati Gumar~\cite{khalifa-etal-2018-morphologically}

\subsection{Annotation Guidelines}
\label{sec:Guidelines}

\textbf{Tokenization}: the text was tokenized 
into sentences, then into raw tokens. 
A raw token can be a word, letter, symbol, 
punctuation mark, or emoji. 
Each token is given a unique identifier. 

\textbf{Token and Spelling Guidelines}: 
A token is the normalized version of the \textit{rawToken}. 
Because there are no standard orthographic spelling 
rules for dialects, 
people typically write words as they pronounce them. 
The same word can be written in many different ways, 
such as \TrAr{اللي} (the one) and \TrAr{الي}, 
\TrAr{هذول} (those ones) 
and \TrAr{هذولا}, \TrAr{مكو} (there is nothing) 
and \TrAr{ماكو} 
or \TrAr{شوية} (a few) and \TrAr{شويا}. 

Also, people sometimes stress certain letters by 
repeating them, such as \TrAr{يسسسسس} (yes), 
\TrAr{ههههههه} (lol), and \TrAr{شووووو} (what). 

In addition, unintentional typos and 
spelling mistakes are more likely to occur in 
social media content. 
More importantly, we noticed that people tend 
to concatenate some functional words 
(e.g., prepositions, pronouns, negation particles) 
with words, such as 
\TrAr{لتخافون} (do not be afraid originally \TrAr{ لا تخافون }).

The lack of such standard orthography makes the 
annotation process challenging. 
One solution is to develop a set of orthographic 
rules for each dialect, and rewrite the 
\textit{rawToken} 
according to these rules. 
This solution (called CODA) was used in the annotation 
of the Palestinian, Lebanese, and Emarati corpora. 

We tried to apply this solution to annotate our 
four corpora. 
However, it was not teachable and did not scale to 
large and diverse corpora. 
Since we have many annotators participating in the 
annotation process, it was difficult to teach 
them CODA rules to maintain the consistency of 
their annotations. 
Instead of using CODA rules, 
we used the following simple normalization rules to 
produce \textit{Token}. 

\begin{enumerate}
\item Unintentional typos and spelling mistakes are 
corrected, 
\item Letters repeated more than two times are 
removed, 
\item Odd and rare spellings are corrected, 
if necessary, to follow more common spellings. 
\end{enumerate} 

With these necessary changes, \textit{Token} 
becomes a normalization, rather than a re-spelling, 
of the \textit{rawToken}. 

\textbf{POS Guidelines}: 
We used the exact SAMA POS tagset to specify 
part-of-speech of the stem of the \textit{Token}. 

\textbf{Segmentation and Affixes Guidelines}: 
Each token is segmented into prefix(es), stem, and suffix(es). 
Each prefix (and suffix) is tagged with its POS. 
Multiple prefixes and suffixes are combined with “+” 
(See Figure~\ref{fig:annotation_example}).  
To maintain compatibility with the tagset of SAMA 
affixes and related corpora, we used the SAMA 
POS affixes, in addition to some dialect-specific 
categories that we discovered and  introduced during 
the annotation process. 

As noted earlier, unlike MSA, some functional words 
(e.g., prepositions, pronouns, negation particles) 
are concatenated with words in dialectal text. 
For example, 
in the word \TrAr{مايعرف} (he does not know), 
the \TrAr{ما} prefix plays the role of a negation 
particle, which we annotated as a prefix. 
Concatenating such different functional words as 
prefixes and suffixes yields a large number of 
prefixes-stem-suffixes combinations indeed. 
See Section~\ref{sec:stats} for statistics. 

\textbf{Lemma Guidelines}: 
Every token is linked with an MSA lemma. 
We used MSA lemmas from the SAMA database. 
In case a lemma is not found in SAMA, 
we used lemmas found in Birzeit's Lexicographic 
database~\cite{JA19} and Arabic Ontology~\cite{J21}; 
otherwise, we introduced a new lemma.
We note here that for dialectal lemmas, 
we added the following. 
\begin{itemize} 
\item Glosses (i.e., senses in Arabic), which is important for word sense disambiguation~\cite{HJ21}, and 
\item Equivalent lemmas in MSA, which is important to link them as synonyms~\cite{GJJB23}.
\end{itemize} 

\textbf{Gloss Guidelines}: 
This is an informal semantic annotation in English. 
By default, we used the glosses of the SAMA Lemmas, 
and edited them if needed. 

\begin{figure*}[t!]
\centering
\includegraphics[width=0.9\textwidth]{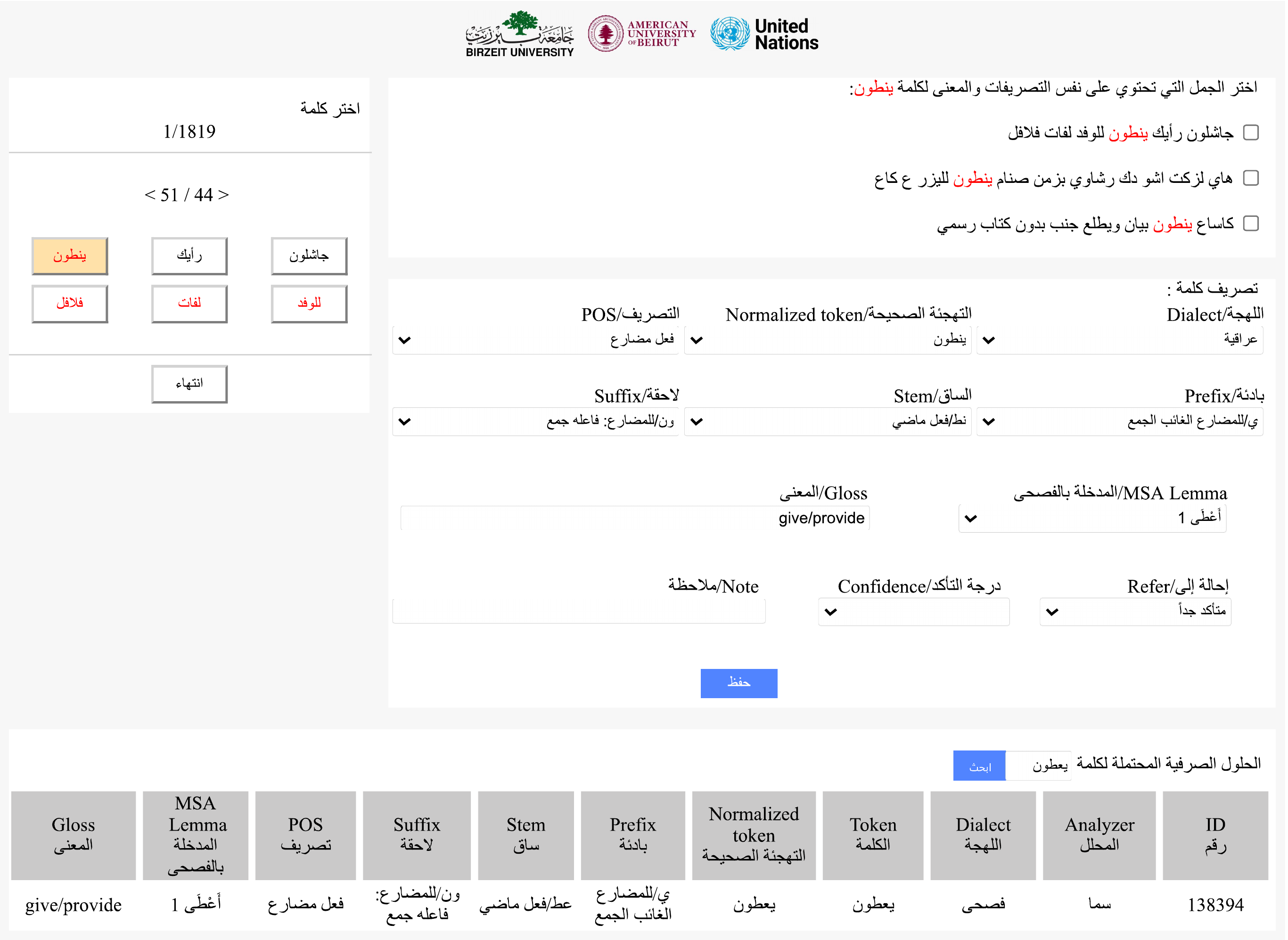}
\caption{Snapshot of \adat in action}
 \label{fig:adat_snaphot} 
\end{figure*}

\subsection{Annotation Methodology}
Each token was annotated manually by native 
speakers using \adat.
\adat supports the annotation guidelines described 
earlier. 
We recruited 35 native speakers and trained them 
on the annotation process and on the use of \adat.

A separate team was established for each dialect, 
led by an expert native speaker. 
The training phase was divided into two steps. 

First, we conducted an online workshop for about 
15 hours. The workshop explained 
the annotation process and the tagsets. 
It included an assisted annotation of a text 
with 200 tokens. 

Second, each annotator was given a corpus of 
$1,000$ words to annotate. 
This served as a quiz set. 
We evaluated the quality of the annotations and 
only the annotators with good quality were recruited. 
The 35 recruited annotators were paid a 
fair rate per hour based on their living locality
(between 5\$ and 10\$ per hour). 
A discussion channel was also created to enable the 
annotators to discuss, post questions, and consult the 
leader on specific issues. 

Each of the four corpora was uploaded to the tool 
and divided into tasks. 
Each task was assigned to an annotator to carry out, 
as described in Section~\ref{sec:adat}. 
The annotation of the four dialects spanned over 
two years.

\subsection{The Annotation Tool}
\label{Adat}
\label{sec:adat}

To guide and speed up the annotation process, 
we developed \adat (see a screenshot in Figure 2), 
which we designed to support collaborative 
annotations. 

At the start, \adat shows the task. 
Each task is a set of words that need to be annotated
and that belong to a specific context 
(same tweet, post, comment).
When the annotator clicks on a word $w$, \adat
retrieves the contexts (sentences) containing the 
$w$. 
The annotator selects those contexts where $w$ appears
with the same semantics as in the original context. 
Annotatint $w$ once, covers annotating $w$ for all the 
selected contexts. 

\adat also displays a set of possible 
morphological solutions for $w$, shown at the 
bottom of the annotation panel.
\adat retrieves these solutions from the following
resources. 
\begin{enumerate}
\item SAMA Database,
\item Curras annotations and 
\item Previous annotations by other annotators using \adat. 
\end{enumerate}

The annotator can tick one or more sentences and 
select the appropriate solution for $w$. 
In case no acceptable solutions were retrieved by 
\adat, fully or partially, 
the annotator can add or edit the annotations. 

After completing the annotation of $w$, 
the annotator must select his/her degree of 
confidence (High, Normal, or Low). 
In case of hesitation about the annotation of 
a certain word, \adat allows the annotators to 
``refer'' the solution to another more experienced
annotator for review. 

The idea of offering the annotators a list 
of suggested annotations helps for 
speed up the annotation process. 
More importantly, it is
critical for minimizing errors and 
maintaining consistency. 
Indeed, we noticed several types of mistakes that 
the annotators made when typing manually, 
which we queried and corrected afterward.

\begin{table*}
\centering
\resizebox{.6\textwidth}{!}{
\begin{tabular}{|lrrrr|}
\hline
\multicolumn{1}{|l|}{Corpus name} & Yemeni & Iraqi & Sudanese & Libyan \\ \hline
Tweets                  & 38,819    & 3,326    & 3,000       & 3,053     \\
Token                             & 1,098,222 & 45,881   & 52,616     & 51,686    \\
Unique Tokens                     & 136,801   & 17,812   & 18,242      & 18,556    \\
Unique Lemma                      & 45,085    & 9,405    & 10,553      & 10,279    \\
Nouns                      & 627,926    & 26,523    & 28,553       & 27,761     \\
Verbs                      & 178,383   & 8,371   & 9,249      & 9,827    \\
Functional words           & 270,844     & 10,242      & 13,448         & 13,239       \\ 
Digit                      & 3,280    & 101    & 7       & 145     \\
Others (e.g., Foreign words)   & 26,643    & 729    & 1,366       & 902     \\
\hline
\end{tabular}
}
\caption{Statistics about the corpus}
\end{table*}

\subsection{Corpus Statistics}
\label{sec:stats} 

\corpora contains more than 1.2 million 
fully annotated tokens, 
represented by 48K documents (tweets, posts and comments)
collected from Yemeni, Iraqi, Sudanese, and Libyan dialects. 
Table~\ref{t:detais} details the number of documents, 
tokens, and lemmas in each of the corpora. 
It also contains the number of unique tokens, 
lemmas, nouns, verbs, digits, functionals, and other
tokens in \corpora.


\section{Conclusion and future work}
\label{sec:conclusion} 
In this paper, we presented 
four morphologically-annotated corpora 
for low-resourced Arabic dialects, 
which consisted of more than $1.2$ million tokens. 
Each word in the four corpora was annotated 
with several morphological features. 
The annotation process was carried out by $35$
native speakers of the target dialects who 
were trained on a set of guidelines and on how 
to use \adat. 
We developed \adat to assist the annotators and to 
ensure compatibility with SAMA and Curras tagsets. 
\adat will be provided as an open sourced contribution
and the four corpora are also available online 
\url{https.portal.sina.birzeit.edu/curras/}. 

We plan to evaluate the four corpora using 
the metrics of the inter-annotator agreement. 
We also plan to refine them and build specialized 
lexicons for the four dialects including
normalizing and unifying subsets of the POS tagets.
We will enrich these lexicons with senses for 
each lemma. 
Last but not least, we plan to use the four 
corpora to extend Wojood~\cite{JKG22} 
by annotating the corpora for 
Named Entity Recognition, 
similar to what we did with Curras and Baldi.

\section*{Acknowledgements}
We would like to thank the 35 annotators and 
the Aklama team who carried out the annotations of 
the four dialects. 
We also acknowledge efforts made by 
students at Birzeit University in reviewing 
and correcting annotations.  
We would like to thank Rayan Dankar in helping 
us develop the first version of the Adat tool, 
as well as for the technical support we received 
from Archetlabs developers.

\bibliography{anthology,custom,references,MyReferences,:}
\bibliographystyle{acl_natbib}
\end{document}